\documentclass{article}

\usepackage{arxiv}

\usepackage[utf8]{inputenc} 
\usepackage[T1]{fontenc}    
\usepackage{hyperref}       
\usepackage{url}            
\usepackage{booktabs}       
\usepackage{amsfonts}       
\usepackage{nicefrac}       
\usepackage{microtype}      
\usepackage{lipsum}
\usepackage{graphicx}

\usepackage{subcaption}

\usepackage{multirow}
\usepackage[table,xcdraw]{xcolor}

\usepackage[ruled,vlined]{algorithm2e}

\usepackage{amsmath}
\usepackage{amssymb}
\usepackage{mathtools}
\usepackage{amsthm}

\definecolor{commentcolor}{RGB}{110,154,155}   
\newcommand{\PyComment}[1]{\ttfamily\textcolor{commentcolor}{\# #1}}  
\newcommand{\PyCode}[1]{\ttfamily\textcolor{black}{#1}} 

\title{One Leaf Reveals the Season: Occlusion-Based Contrastive Learning with Semantic-Aware Views for Efficient Visual Representation}

\author{
  Xiaoyu Yang \\
  Shanghai AI Lab \\
  Shanghai \\
  China \\
  \texttt{xiaoyuyang386@gmail.com} \\
  \And
  Lijian Xu\\
  Shanghai AI Lab \\
  Shanghai \\
  China \\
  \And
  Hongsheng Li \\
  Department of Electronic Engineering \\
  the Chinese University of Hong Kong \\ 
  Hongkong \\
  China \\
  \AND
  Shaoting Zhang \\
  Shanghai AI Lab \\
  Shanghai \\
  China \\
}

\begin{document}
\maketitle

\begin{abstract}
  This paper proposes a scalable and straightforward pre-training paradigm for efficient visual conceptual representation called occluded image contrastive learning (OCL). Our OCL approach is simple: we randomly mask patches to generate different views within an image and contrast them among a mini-batch of images. The core idea behind OCL consists of two designs. First, masked tokens have the potential to significantly diminish the conceptual redundancy inherent in images, and create distinct views with substantial fine-grained differences on the semantic concept level instead of the instance level. Second, contrastive learning is adept at extracting high-level semantic conceptual features during the pre-training, circumventing the high-frequency interference and additional costs associated with image reconstruction. Importantly, OCL learns highly semantic conceptual representations efficiently without relying on hand-crafted data augmentations or additional auxiliary modules. Empirically, OCL demonstrates high scalability with Vision Transformers, as the ViT-L/16 can complete pre-training in 133 hours using only 4 A100 GPUs, achieving 85.8\% accuracy in downstream fine-tuning tasks. Code is available at \url{https://anonymous.4open.science/r/OLRS/}.
\end{abstract}


\section{Introduction}
\label{sec:intro}

Self-supervised learning (SSL) is a key approach for building world models, especially for pre-training vision models \cite{chenSimpleFrameworkContrastive2020,heMaskedAutoencodersAre2022,baoBEiTBERTPreTraining2021,radfordLearningTransferableVisual2021,wangDropPosPreTrainingVision2023,yang2024enhancingvisualgroundinggeneralization}. Its strength lies in learning versatile visual representations without relying on human annotations.
Currently, the two main paradigms in visual SSL are Masked Image Modeling (MIM) \cite{heMaskedAutoencodersAre2022, wangDropPosPreTrainingVision2023, kongUnderstandingMaskedAutoencoders2023,zhangHowMaskMatters2022, guptaSiameseMaskedAutoencoders2023} and Contrastive Learning (CL) \cite{chenSimpleFrameworkContrastive2020}. Both have shown strong scalability, particularly for Vision Transformers (ViTs) \cite{dosovitskiyImageWorth16x162021}.

Despite the success of existing methods, both MIM and CL struggle to achieve efficient visual representation. MIM focuses heavily on pixel-level reconstruction, which often prioritizes local details over high-level semantic concepts. Similarly, CL suffers from conceptual redundancy, where transformed images may lack meaningful differences.
This raises a critical question: 
\textit{How can we bridge the gap between efficient visual representation and effective conceptual pre-training, overcoming the limitations of current MIM and CL approaches?
}

\begin{figure*}[htbp]
    \centering
    \begin{subfigure}[t]{0.18\textwidth}
        \centering
        \includegraphics[height=.22\textheight]{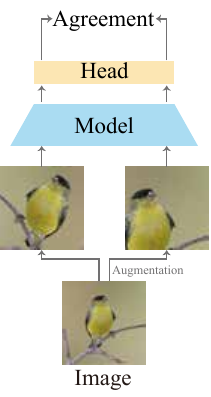}
        \caption{CL}
        \label{fig:cl}
    \end{subfigure}
    \begin{subfigure}[t]{0.2\textwidth}
        \centering
        \includegraphics[height=.22\textheight]{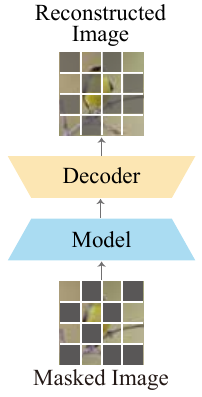}
        \caption{MIM}
        \label{fig:mim}
    \end{subfigure}
    \begin{subfigure}[t]{0.6\textwidth}
        \centering
        \includegraphics[height=.21\textheight]{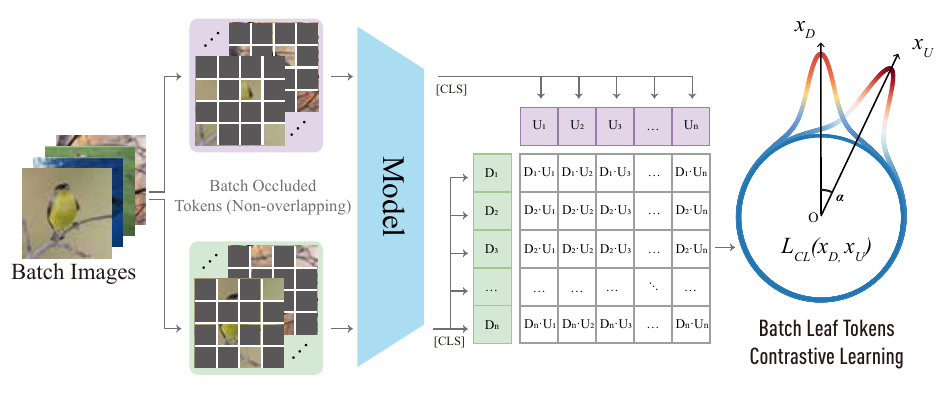}
        \caption{Our occluded image contrastive learning}
        \label{fig:mcl}
    \end{subfigure}    
        \caption{\textbf{Comparison between different pre-training paradigms.} The Model in blue is the pre-training model, and the orange modules indicate auxiliary modules. (a) Contrastive Learning (CL) endeavours to maximize the agreement between different views of an image. (b) Masked Image Modeling (MIM) aims to restore masked image patches. (c) Our occluded image contrastive learning: Through non-overlapping occluding, distinct tokens within an image are categorized as intraclass, while across-image tokens within a batch are viewed as interclass. Our objective is to enhance intraclass compactness and interclass separability through a contrastive learning approach. Just as a single leaf can tell the coming of autumn, we believe that a small area of the image contains the majority of the meaning of the entire image. 
        }
    \label{fig:model}
\end{figure*}

To address this question, we first revisit the two main pre-training paradigms: MIM and CL. MIM learns visual representations by reconstructing masked image patches (see Fig.\ref{fig:mim}). Notable examples include BEiT \cite{baoBEiTBERTPreTraining2021} and MAE \cite{heMaskedAutoencodersAre2022}. MAE highlights that images contain significant semantic redundancy, meaning only a basic understanding of objects and scenes is needed to predict missing patches from their surroundings.
However, pixel-level reconstruction is too detailed for pre-training vision models. It focuses excessively on high-frequency details and local features, conflicting with the goal of pre-training: learning high-level semantic concepts. While this detailed task helps models learn visual representations, it comes at the cost of pre-training efficiency.

In CL, the core idea is simple: maximize agreement between different views of the same image (see Fig.\ref{fig:cl}). Popular methods like SimCLR \cite{chenSimpleFrameworkContrastive2020,chenBigSelfSupervisedModels2020}, MoCo v3 \cite{chenEmpiricalStudyTraining2021a}, and DINO \cite{caronEmergingPropertiesSelfSupervised2021} use complex pre-processing and auxiliary networks to create distinct views of an image. The main challenge lies in optimizing the agreement between these views.
For CL to work well, distinct views are essential. However, this is difficult because images often have conceptual redundancy. Additionally, the dependency on large-scale batches to generate sufficiently diverse negative samples imposes stringent computational constraints,  notably escalating computational and time consumption overhead during training.


Driven by this analysis, we found that these two paradigms can complement each other: masked tokens have the potential to significantly diminish the conceptual redundancy inherent in images, whereas contrastive learning is adept at extracting high-level semantic features during the pre-training phase. Thus, we present a novel and straightforward paradigm for self-supervised visual representation learning: occluded image contrastive learning (OCL). OCL tackles the above issues systematically: I) Masked image tokens offer diverse views of a single image with substantial fine-grained conceptual differences. II) Contrastive learning enables pre-training to concentrate exclusively on the high-level semantic information contained within images while disregarding high-frequency redundancies. III) The proposed paradigm obviates the need for auxiliary modules and expedites the efficient extraction of model features.

OCL has a particularly simple and straightforward workflow, as presented in Fig.\ref{fig:mcl}. Here is how it works: Firstly, we mask a batch of images with a high rate, dividing visible patches within one image into two non-overlapping groups. In succession, the pre-train model extracts the features of these two groups of batch image tokens, respectively. Subsequently, contrastive learning is employed to predict the correct pairings for a batch of visible image tokens. Positive samples are different visible tokens in the same image, while negative samples are from different images of the mini-batch. Finally, inspired by T-distributed classifier  \cite{yangTdistributedSphericalFeature2023, yang2024adaptingmultimodallargelanguage}, we use the T-distributed spherical loss to constrain the inter-class margins in the pre-training. 
Comprehensive experiments demonstrate the scalability and efficacy of our approaches, where ViT-L/16 can complete pre-training in 133 hours using only 4 A100 GPUs and attain an 85.8\% top-1 accuracy in fine-tuning classification. In particular, our model stands out from other pre-training methods as it operates without the need for auxiliary modules or hand-crafted data augmentation to generate diverse views.

In summary, our paper mainly makes the following contributions:
\begin{enumerate} 

    \item We endeavour to explore an alternative of using masked images to create diverse views with fine-grained conceptual differences for contrastive learning. By forgoing the conventional approach of employing instance-level hand-crafted data augmentation to generate distinct views, OCL diminishes the conceptual redundancy inherent in images and improves efficiency.

    \item Our approach eschews the reconstruction of masked images in favour of leveraging contrastive loss to steer the entire model. Independently of additional auxiliary modules, OCL is adept at extracting high-level semantic concept features from images more efficiently.

    \item Extensive experiments are conducted to verify the efficiency and scaling capability of our method. ViT-L/16 can complete pre-training in 133 hours using only 4 A100 GPUs with 85.8\% accuracy in fine-tuning. Additionally, we have structured ablation experiments to delve into the implications of different configurations within OCL, with a particular focus on the need of the MLP head in contrastive learning.
\end{enumerate}

\section{Methodology}

\subsection{Architecture}

With an input image $ x_{i}  \in  \mathbb{R}^{H\times W \times C} $, it is reshaped into a sequence of 2D patches $x_{p}\in \mathbb{R}^{N\times (P^{2}\cdot C)}$, where $(H, W)$ denotes the original image resolution, $C$ is the number of channels, $P$ represents the patch size, and $N = HW/P^{2}$ indicates the number of patches. Subsequently, a linear projection is employed on $x_{p}$ to transform it into $D$ dimensions, yielding patch embeddings $x\in\mathbb{R}^{N\times D}$. 
Following the MoCo v3 \cite{chenEmpiricalStudyTraining2021a}, the linear projection is initialized using the Xavier uniform method and remains fixed throughout pre-training to mitigate potential instability in ViT caused by the large batch size. 
Thereafter, fixed position embeddings $p\in\mathbb{R}^{(N+1) \times D}$ are incorporated into the patch embeddings to preserve positional information, employing sinusoidal positional encoding. 

After random masking patches, visible patches that retain the original image position information are divided into two non-overleaping groups: $x_{U}\in\mathbb{R}^{n\times D}$ and $x_{L} \in \mathbb{R}^{{n\times D}}$, where $n$ denotes the number of visible patches. Each group adds an independent \textbf{[CLS]} token $x_{cls}\in\mathbb{R}^{D}$ to aggregate the information of each group. Moreover, \textbf{[CLS]} tokens of each group will add the position embeddings $p_{0}$. In succession, ViT \cite{dosovitskiyImageWorth16x162021} is utilized as our encoder. $z_{U} = [x^{U}_{cls};x_{U}] \oplus p$ and $z_{L} = [x^{L}_{cls};x_{L}] \oplus p$ are the input of pre-training ViT, where $\oplus$ denotes element-wise plus, and $f(\cdot)$ represents the ViT. Consequently, image feature tokens extracted by ViT are symbolized as $y = f(z)$, where $y\in\mathbb{R}^{(n+1)\times D}$. Furthermore, we only preserve \textbf{[CLS]} tokens $y_{cls}\in\mathbf{R}^{D}$ of both groups for contrastive learning, as it aggregates the high-level semantic information of each image.
It is noteworthy that, unlike existing methodologies, we abstain from employing auxiliary modules. This allows our model to more easily extract image features and reduce the consumption of computing resources.

\begin{figure}[htb]
    \centering
    \includegraphics[width=.48\textwidth]{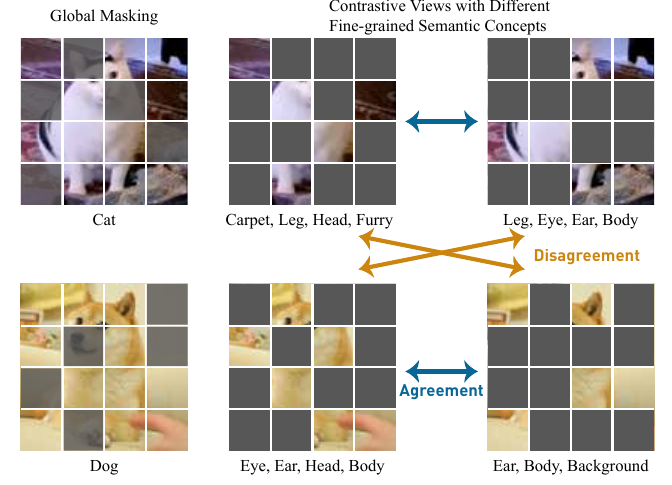}
    \caption{\textbf{A toy example of masked images for conceptual contrastive learning.} The low global masking ratio aids the model in capturing comprehensive information from the image and understanding the interconnectedness of various concepts within a mini-batch. Beyond that, each contrastive branch has a higher masking ratio, generating diverse views with different semantic concepts for contrastive learning and diminishing conceptual redundancy within the image. 
    }
    \label{fig:toy}
\end{figure}

\subsection{Occluded Image for Conceptual Pre-training}

Despite the strides made in instance-wise contrastive learning  \cite{chenSimpleFrameworkContrastive2020,caronEmergingPropertiesSelfSupervised2021}, the differences between various views primarily manifest at the pixel level, instead of semantic concept disparities. In this context, a patch is abstracted as a concept containing fine-grained semantics. The concepts within an image exhibit distinct conceptual characteristics, yet they are all interconnected with the overall meaning of the image, albeit to varying extents. Thus, masked images are used to create views that contain semantic distinctions of concepts, and mitigate the conceptual redundancy present in images, as presented in Fig. \ref{fig:toy}.

We apply the masking strategy of random sampling, the same as the MAE, which samples random patches without replacement following the uniform distribution. Beyond that, a low masking ratio  (e.g. 30\%) is implemented overall, but leads to a higher masking ratio (e.g. 65\%) for individual contrastive branches, as illustrated in Fig. \ref{fig:toy}. 
The low global masking rate helps the model capture the overall image structure and the relationships between different concepts within a mini-batch. During each forward pass, the model only sees a small portion of the visible tokens. This creates diverse views with fine-grained semantic differences, making the task more challenging and reducing the risk of the model relying on simple patterns or redundant information.
The high masking ratio for each contrastive branch ensures that the task cannot be solved by simply extrapolating from basic image transformations. This reduces conceptual redundancy and forces the model to focus on localized features. Additionally, by training on only a few parts of the image, the masking strategy also improves scalability and efficiency, making it easier to handle the large batch sizes required for contrastive learning.


To divide visible patches into two separate, non-overlapping groups, we use a random sampling strategy similar to the masking approach described earlier. This helps avoid potential biases and ensures that the central positions of the visible patches are consistent across both groups.
Meanwhile, non-overlapping patches present a challenging scenario, impeding the model from relying solely on analogous patches for inference. 





\subsection{Efficient Contrastive Learning}

We randomly select a mini-batch of $B$ instances and establish the contrastive prediction task on the visible token pairs extracted from this mini-batch, yielding a total of $2\times B$ data points. 
Feature tokens from different groups within the same image are treated as positive pairs, while tokens from different images in the same mini-batch are treated as negative pairs. 
We concurrently maximise the similarity of $B$ positive pairs while minimizing the similarity of $B^{2}-B$ negative examples to drive the network.

In terms of similarity computation, we introduce T-distributed spherical (T-SP) metric  \cite{yangTdistributedSphericalFeature2023,kobayashiTvMFSimilarityRegularizing2021} to significantly promote the intraclass compactness and interclass separability of features. Given \textbf{[CLS]} tokens $y_{i}^{U}\in\mathbf{R}^{D}$ and $y_{j}^{L}\in\mathbf{R}^{D}$ of both non-overlapping groups, the cosine distances between $y_{i}^{U}$ and $y_{j}^{L}$ are:
\begin{equation}
    cos_{LU}(y_{i}^{U},y_{j}^{L}) = \frac{{y_{i}^{U}}^{T}y_{j}^{L}}{|y_{i}^{U}||y_{j}^{L}|}
\end{equation}
and the T-SP similarity is defined as follows:
\begin{equation}
    \text{sim}_{tsp}(y_{i}^{U},y_{j}^{L}) = 0.5\times\frac{1 + cos_{LU}}{1 + (1-cos_{LU})*\kappa}
\end{equation}
where $\kappa\geq0$ denotes the concentration hyperparameter of T-SP metric. As $\kappa$ decreases, the similarity function becomes more condensed, where only two tokens in close proximity are deemed positive examples. Besides, we add a trainable temputare parameter $\tau$ to effectively scale the different samples. Thus, the loss function for a positive pair is:
\begin{equation}
        \mathcal{L}(y_{i}^{U},y_{j}^{L}) 
        = -\log{
            \frac{\exp{(\text{sim}_{tsp}(y_{i}^{U},y_{j}^{L})\times\tau)}}{
                \sum_{k=1}^{2B}{
                    \mathbb{1}_{ \left[ k \neq i \right] } 
                    \exp{(\text{sim}_{tsp}(y_{i}^{U},y_{k}^{L})\times\tau)}
                }
            }
        }    
\end{equation}
where $\mathbb{1}_{ \left[ k \neq j \right] } \in \{0, 1\}$ is an indicator function evaluating to 1 if $k\neq i$. Finally, inspired by CLIP  \cite{radfordLearningTransferableVisual2021}, we optimize a symmetric loss over these similarity scores within a mini-batch exhibited in Algorithm \ref{algo:your-algo}.






\subsection{Simple Implementation}

\begin{algorithm}[h]
\SetAlgoLined

    \PyComment{x[B,N,D] - patch embeddings } \\
    \PyCode{} \\    
    \PyComment{mask image and split into two non-overleaping groups} \\
    \PyComment{x [B$\times$2, n, L]} \\
    \PyCode{x = masking(x, ratio = 0.3)} \\
    \PyComment{extract [CLS] token of both groups using pre-training model} \\    
    \PyCode{x = model(x)} \PyComment{[B$\times$2, 1, L]} \\
    \PyCode{x = x.reshape(-1, 2, x.shape[-1])} \\
    \PyComment{L2 normalize the [CLS] token of each group } \\    
    \PyCode{m, n = x[:,0], x[:,1]} \PyComment{[B, L]}\\
    \PyCode{m = m/m.norm(dim=-1,keepdim=True)} \\
    \PyCode{n = n/n.norm(dim=-1,keepdim=True)} \\
    \PyComment{compute the scaled pairwise T-SP similarities} \\ 
    \PyCode{sim\_mn = compute\_tSP(m @ n.T)} \\
    \PyCode{sim\_nm = compute\_tSP(n @ m.T)} \\
    \PyComment{symmetric loss function} \\ 
    \PyCode{labels = torch.arange(B)} \\
    \PyCode{loss=(F.cross\_entropy(sim\_mn,labels)}
    \PyCode{+F.cross\_entropy(sim\_nm,labels))/2} \\
\caption{Pytorch-like pseudo-code for the core of an implement of OCL}
\label{algo:your-algo}
\end{algorithm}

The implementation of our OCL pre-training is efficient and involves minimal specialized operations. As pseudo code depicted in Algorithm \ref{algo:your-algo}, we make only minor modifications based on the MAE code, mainly involving the process subsequent to the acquisition of image feature embeddings from the encoder. First, we randomly mask a subset of embedded patch tokens with a low masking ratio. In succession, listed tokens are shuffled randomly and divided into two non-overlapping groups. Following MAE \cite{heMaskedAutoencodersAre2022}, within positional and $\text{[CLS]}$ embeddings, lists of tokens are encoded by the ViT. It is noteworthy that we obtain the encoded $\text{[CLS]}$ token from the ViT directly, without the incorporation of additional auxiliary modules, even a linear head or lightweight decoder.

Subsequently, the obtained $\text{[CLS]}$ tokens within each group are L2 normalized, and then the T-SP metric is applied to calculate the similarity between tokens from each group. Finally, a simple cross-entropy loss is calculated symmetrically to drive model training, enhancing the intraclass conceptual compactness within an image and the interclass semantic separability across images.


\section{Experiments}


\begin{table}[htbp]
\centering
\begin{tabular}{@{}ccccc@{}}
\toprule
Model     & Blocks & Dim  & Heads & Params \\ \midrule
ViT-B/16  & 12     & 768  & 12    & 86M    \\
ViT-L/16  & 24     & 1024 & 16    & 304M   \\
ViT-h/16  & 32     & 1280 & 16    & 632M   \\ \bottomrule
\end{tabular}
\caption{\textbf{Configurations of Vision Transformer \cite{dosovitskiyImageWorth16x162021} models in our experiments.} Block denotes the number of transformer blocks, with dim representing the input/output channel dimension of all blocks. Heads are the number of heads in multi-head attention modules. We also provide the parameter sizes of different models.}
\label{Tab.Configure}
\end{table}

We perform self-supervised pre-training on the ImageNet-1K \cite{russakovskyImagenetLargeScale2015} dataset with the resolution of 224$\times$224. By default, ViT-B/16 and ViT-L/16 \cite{dosovitskiyImageWorth16x162021} are leveraged as the backbone architecture with 800 epochs for pre-training and 40 epochs for warm-up. ViT-B/16 applies the overall masked ratio of 0.3, while ViT-L/16 is set to 0.4. The initial base learning rate is $1.5 \times 10^{-4}$. Like other contrastive learning \cite{chenSimpleFrameworkContrastive2020,radfordLearningTransferableVisual2021}, our method relies on a large effective batch size: 9,600 for ViT-B/16 and 2,048 for ViT-L/16. 
Besides, $\kappa$ is set to 64 in the T-SP metric for computing similarity by default.
Our implementation is based on MAE \cite{heMaskedAutoencodersAre2022}, with further details provided in the Supplementary Material.

In terms of supervised validation, OCL is evaluated through end-to-end fine-tuning and linear probing on the ImageNet-1k dataset for classification with 100 epochs for ViT-B/16 and 50 for ViT-L/16, following common practices \cite{caronEmergingPropertiesSelfSupervised2021,heMaskedAutoencodersAre2022,wangDropPosPreTrainingVision2023}. Top-1 accuracy is utilized to verify the performance of different methods.

\subsection{Scalability}

\begin{figure}[htbp]
    \centering
    \includegraphics[width=.48\textwidth]{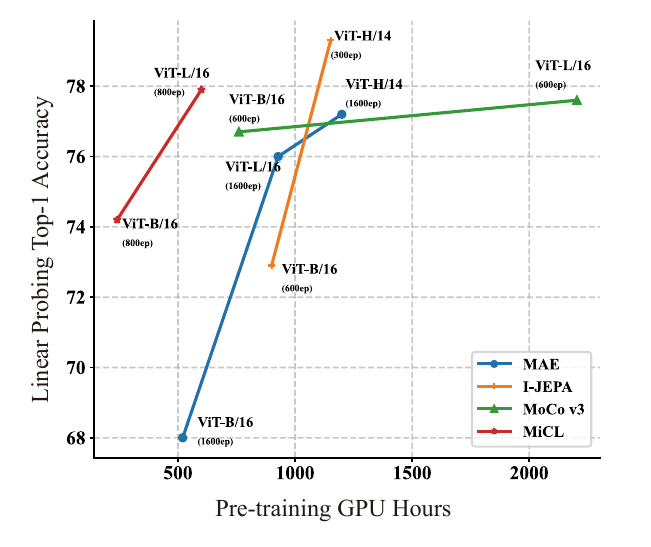}
    \caption{\textbf{Efficiency and Scaling.} MAE \cite{heMaskedAutoencodersAre2022}, I-JEPA \cite{assranSelfSupervisedLearningImages2023} and MoCo v3 \cite{chenEmpiricalStudyTraining2021a} are opted for comparison. All methods are evaluated by linear probing with Top-1 accuracy (Acc) as the metric, and the pre-training GPU time with A100 hour as the indicator. The pre-training epochs (denoted as ep) and model architecture are also exhibited.
    }
    \label{fig:eff}
\end{figure}

To demonstrate the scalability of our OCL for efficient conceptual pre-training, we access the efficiency and scaling of our model in Fig.\ref{fig:eff}.  It illustrates the pre-training hours related to various model sizes for different methods, with linear probing accuracy.
OCL is highly scalable compared to previous methods, requiring less computational resources while achieving strong results and without relying on handcrafted data augmentations. Unlike reconstruction-based approaches like MAE and I-JEPA, OCL needs fewer training epochs and avoids the need for pixel-level reconstruction, significantly improving training speed.
In contrast to contrastive learning methods such as MoCo v3, which depend on handcrafted augmentations and complex architectures to generate and process multiple image views, OCL eliminates the need for auxiliary modules like momentum encoders. This simplicity makes OCL’s framework more efficient and accelerates pre-training.
For example, when scaling up from ViT-B/16 to ViT-L/16, OCL requires far less additional pre-training time compared to MoCo v3.

\textbf{Scaling model size.} Moreover, our model leverages a scalable model size, resulting in more substantial performance enhancements with the larger model as illustrated in Fig.\ref{fig:eff}. Compared to ViT-B/16, OCL achieves nearly 4\% improvement in linear probing with ViT-L/16, surpassing MoCo v3. It implies that we can efficiently train larger models to achieve better performance, within an acceptable timeframe.

\subsection{Ablation Studies}

\subsubsection{Masked Ratio}

\begin{table}[htbp]
\begin{tabular}{@{}ccccc@{}}
\toprule
\begin{tabular}[c]{@{}c@{}}Overall\\      Masked Ratio\end{tabular} & \begin{tabular}[c]{@{}c@{}}Forward \\      Visible Ratio\end{tabular} & Eff. Bsz. & LIN  & FT   \\ \midrule
\multicolumn{5}{l}{\textit{ViT-L/16}}                                                                                                                                 \\
0.2                                                                 & 0.4                                                                   & 1,024     & 74.2 & 84.4 \\
0.4                                                                 & 0.3                                                                   & 2,048     & 77.9 & 85.8 \\
0.8                                                                 & 0.1                                                                   & 7,200     & 69.1 & 83.0 \\ \midrule
\multicolumn{5}{l}{\textit{ViT-B/16}}                                                                                                                                 \\
0.3                                                                 & 0.35                                                                  & 9,600     & 74.2 & 83.4 \\
0.6                                                                 & 0.2                                                                   & 9,600     & 71.0 & 82.8 \\
0.8                                                                 & 0.1                                                                   & 12,800    & 65.7 & 81.6 \\ \bottomrule
\end{tabular}
\caption{\textbf{Ablation evaluation experiments on masked ratio.} The results are based on the ImageNet-1K with the Vit-B/16 and ViT-L/16. All methods are evaluated by linear probing (LIN) and fine-tuning (FT). We provide the visible ratio of each branch for contrastive learning according to the masked ratio. Correspondingly, we give different effective batch sizes (Eff. Bsz.) related to the overall masked ratio. The resolution of images is fixed to 224×224. Top-1 accuracy is used as the metric. }
\label{Tab.Ratio}
\end{table}

Firstly, we conduct ablation experiments to discuss the impact of the overall masked ratio on the performance of conceptual pre-training, revealed in Table \ref{Tab.Ratio}. To ensure clarity in presentation, we offer the visible ratio of each contrastive branch within a single forward pass of the mini-batch, which is a crucial factor for the cross-similarity within our OCL framework. Likewise, the effective batch size is intertwined with the scalable masked ratios, with larger batch sizes showcasing performance enhancements for the model, as verified in Section \ref{sec:bsz}. Thus, we attribute this improvement to the masked ratio.

Regarding large models such as ViT-L/16 in Table \ref{Tab.Ratio}, the increment of the overall masked ratio from 0.2 to 0.4 could significantly diminish conceptual redundancy and increase the effective batch size, thereby resulting in performance improvement. Nevertheless, with a continued rise in the masked ratio, the visible patches of images in one forward process of the model diminish incrementally. Despite increases in batch size, the precise extraction of conceptual information from the images becomes compromised, resulting in degraded performance. Furthermore, as displayed in ViT-B/16 of Table \ref{Tab.Ratio}, once the batch size surpasses its threshold, an excessively high masked ratio can impair the model performance, transitioning from reducing conceptual redundancy to damaging essential semantic information.

\subsubsection{Large Batch Size for Effective Contrastive Representation}
\label{sec:bsz}

\begin{table}[htbp]
\centering
\begin{tabular}{@{}cccc@{}}
\toprule
\begin{tabular}[c]{@{}c@{}} Effective \\ Batch Size \end{tabular}  & LIN  & FT   & \begin{tabular}[c]{@{}c@{}}Pre-training \\ Hour\end{tabular} \\ \midrule
2,048     & 77.9 & 85.8 & 533   \\
4,096     & 77.7 & 85.7 & 533   \\
1,800     & 77.4 & 85.7 & 559   \\
1,024     & 74.4 & 84.9 & 586   \\
512       & 71.6 & 84.6 & 613   \\ \bottomrule
\end{tabular}
\caption{\textbf{Ablation evaluation experiments on batch size.} The results are based on the ImageNet-1K with the ViT-L/16. All methods are evaluated by pre-training hours, linear probing (LIN) and fine-tuning (FT).  Meanwhile, pre-training hours on A100 are provided. The resolution of images is fixed to 224×224. Top-1 accuracy is used as the metric.}
\label{Tab.Bsz}
\end{table}

ViT models \cite{dosovitskiyImageWorth16x162021} are inherently computationally intensive, and training with large batches is a preferred strategy for handling large ViT models. Moreover, a sizable batch size is advantageous for achieving accuracy in contemporary self-supervised learning techniques. In particular, concerning contrastive learning methodologies that heavily lean on large batch sizes, ablation experiments are conducted to ascertain the influence of batch size on our occluded image contrastive learning approach, as shown in Table \ref{Tab.Bsz}. 
The utilization of the ViT-L/16 model for validation reveals that a larger effective batch size correlates with improved model performance and efficiency, aligning with the consensus within the community \cite{goyalAccurateLargeMinibatch2018,youLargeBatchOptimization2019}. 
We attribute this to the statistical advantage of expanded negative sample pools, which achieve broader coverage of the latent feature space. Such comprehensive sampling sharpens inter-class separation by refining decision boundaries during contrastive optimization.
However, limited by computational resources, we are unaware of the maximum capacity of the effective batch size. 

\subsubsection{MLP Head is Not You Need}

\begin{table}[htbp]
\centering
\begin{tabular}{@{}lcccc@{}}
\toprule
MLP Head & \begin{tabular}[c]{@{}c@{}}Pre-training\\      Hours\end{tabular} & Eff. Bsz. & LIN  & FT   \\ \midrule
w/o      & 533 & 2,048     & 77.9 & 85.8 \\
w/o      & 559 & 1,800     & 77.4 & 85.7 \\
2-layer  & 600 & 1,800     & 76.7 & 85.6 \\
3-layer  & 611 & 1,800     & 76.6 & 85.6 \\ \bottomrule
\end{tabular}
\caption{\textbf{Ablation experiments on MLP head.} The results are based on the ImageNet-1K with the ViT-L/16. All methods are evaluated by pre-training hours, linear probing (LIN) and fine-tuning (FT). Correspondingly, we give different effective batch sizes related to the MLP head within the pre-trained model. Besides, pre-training hours on A100 are provided. The resolution of images is fixed to 224×224. Top-1 accuracy is used as the metric.}
\label{Tab.MLP}
\end{table}

In many contrastive learning methods \cite{chenSimpleFrameworkContrastive2020,chenEmpiricalStudyTraining2021a}, the ViT model is typically paired with an MLP head \cite{taudMultilayerPerceptronMLP2018} to assist with the pretext task. The ViT learns semantic features from the image, while the MLP head handles the pretext classification task, enhancing the ViT's learning capabilities, as shown in earlier work.
However, traditional contrastive learning methods rely on handcrafted data augmentations to generate and process multiple image views at the instance level. In contrast, OCL explores an alternative approach: using masked images to create diverse views with fine-grained conceptual differences at the token level. As a result, OCL does not require an MLP head for instance-level classification tasks.

The ablation experiments on the MLP head involve testing different configurations, including a 2-layer MLP, a 3-layer MLP, and no MLP head (see Table \ref{Tab.MLP}). The design of the MLPs follows MoCo v3 \cite{chenEmpiricalStudyTraining2021a}, where the hidden layers of both the 2-layer and 3-layer MLPs are 1024-dimensional and employ the GELU activation function \cite{hendrycksGaussianErrorLinear2023}. The output layers of both MLPs are 512-dimensional and do not use GELU. Additionally, all layers in the MLPs incorporate Batch Normalization (BN) \cite{ioffeBatchNormalizationAccelerating2015}, consistent with the methodology in SimCLR \cite{chenBigSelfSupervisedModels2020}.
Due to GPU memory constraints, the batch size was adjusted to 1,800 for validation. From the fine-tuning results presented in Table \ref{Tab.MLP}, OCL is capable of operating effectively without an MLP head. Unlike traditional contrastive learning approaches, the absence of an MLP head does not degrade the model's performance.
On the other hand, the inclusion of additional auxiliary modules, such as deeper MLP heads, increases both the cost and time required for pre-training. However, the benefits of these modules do not outweigh the performance gains achieved by simply increasing the batch size. 


\subsubsection{Concentration of Similarity Metric $\kappa$}

\begin{table}[htbp]
\centering
\setlength{\tabcolsep}{3mm}{
\begin{tabular}{cccccc}
kappa & 4    & 16   & 32    & 64     & 128 \\ \hline
FT    & 84.8 & 85.1 & 85.6  & 85.8   & 85.5 
\end{tabular}
}
\caption{\textbf{Ablation experiments on concentrate parameter of kappa.} The results are based on the ImageNet-1K with the ViT-L/16. Fine-tuning (FT) results are provided. The resolution of images is fixed to 224×224. Top-1 accuracy is used as the metric.}
\label{Tab.Kappa}
\end{table}

Following the approach of T-SP \cite{yangTdistributedSphericalFeature2023}, we conduct ablation experiments to evaluate how different concentrations of the T-distributed adapter affect the pre-training model’s performance. The results are shown in Table \ref{Tab.Kappa}.
This study involves five specific degrees of concentration, namely $\kappa=4$, $16$, $32$, $64$, and $128$. From the Table \ref{Tab.Kappa}, we can infer that as kappa increases, the pre-training performance of the model improves progressively until reaching $\kappa=64$ with the fine-tuning result of 85.8 under ViT-L/16. 
We explain that as kappa increases, the model must extract more precise semantic concepts from positive samples, creating a more challenging pretext task for pre-training. However, if kappa becomes too large, the model converges too slowly, which negatively impacts the overall effectiveness of the pre-training process.

\subsection{Comparisons with previous results}

\begin{table}[htb]
\centering
\begin{tabular}{@{}lcccccc@{}}
\toprule
\multicolumn{1}{c}{\multirow{2}{*}{Methods}} & \multirow{2}{*}{Aug.} & \multirow{2}{*}{Ep.} & \multicolumn{2}{c}{ViT-B/16} & \multicolumn{2}{c}{ViT-L/16} \\
\multicolumn{1}{c}{}                         &                       &                      & LIN           & FT           & LIN           & FT           \\ \midrule
\multicolumn{7}{l}{\textbf{Masked Image   Modeling}}                                                                                                      \\ \midrule
BEiT                                         & w/o                   & 800                  & -             & 83.2         & -             & 85.2         \\
MAE                                          & w/o                   & 1,600                & 68.0          & 83.6         & 76.0          & 85.9         \\
CAE                                          & w/o                   & 1,600                & 70.4          & 83.9         & 78.1          & 86.3         \\
I-JEPA                                       & w/o                   & 600                  & 72.9          & -            & 77.5          & -            \\ \midrule
\multicolumn{7}{l}{\textbf{Contrastive Learning}}                                                                                                         \\ \midrule
DINO                                         & w/                    & 1,600                & 78.2          & 82.8         & -             & -            \\
MoCo v3                                      & w/                    & 600                  & 76.7          & 83.2         & 77.6          & 84.1         \\ \midrule
\multicolumn{7}{l}{\textbf{Masked Image   Modeling with Contrastive Learning}}                                                                            \\ \midrule
SiameseIM                                    & w/                    & 1,600                & 78.0          & 84.1         & -             & -            \\
ccMIM                                        & w/o                   & 800                  & 68.9          & 84.2         & -             & -            \\
ConMIM                                       & w/                    & 800                  & -             & 85.3         & -             & 86.5         \\
iBOT                                         & w/                    & 1,600                & 79.5          & -            & 81.0          & 84.0         \\ \midrule
OCL                                         & w/o                   & 1,600                & 74.2          & 83.4         & 77.9          & 85.8         \\ \bottomrule
\end{tabular}
\caption{\textbf{Comparsion with previous methods on ImageNet-1K classification.} All methods are evaluated by linear probing (LIN) and fine-tuning (FT). The resolution of images is fixed to 224×224. Aug. indicates the utilization of handcrafted view data augmentation during pre-training. Arch. represents the model architecture, where ViT-B/16 is marked in blue to distinguish different methods. Ep. denotes the epochs of pre-training. Top-1 accuracy (Acc) is used as the metric. }
\label{Tab.Com}
\end{table}

To demonstrate OCL’s ability to learn high-level conceptual representations without relying on handcrafted data augmentations, we compare its performance on linear probing and fine-tuning tasks under pre-training on ImageNet-1k. Importantly, while masked images are used to reduce semantic redundancy, OCL avoids reconstructing masked images. Instead, it uses contrastive loss to guide the network’s learning process. As a result, OCL is categorized as a contrastive learning method.

Table.\ref{Tab.Com} exhibits the performance of our method under the fine-tuning and linear probing for ImageNet-1k classification. Our OCL method has demonstrated outstanding performance in contrastive learning, notably obviating the need for intricate augmentations across multiple views. It validates the practicality and efficacy of masked images for generating diverse views encapsulating distinct fine-grained semantic concepts, which diminishes conceptual redundancy and expedites the conceptual pre-training process. Beyond that, it also stands out the significant competency of contrastive learning in extracting high-level semantic concepts.

Compared to MIM methods, due to the absence of pixel-level image reconstruction in OCL, the training process is finished in 800 epochs, in contrast to the 1600 epochs required for MAE \cite{heMaskedAutoencodersAre2022}, CAE \cite{chenContextAutoencoderSelfsupervised2024}. It corroborates the contribution of our methodology from the perspective of efficiency and high-level semantic concept extraction. Moreover, our OCL method is superior to BEiT \cite{baoBEiTBERTPreTraining2021} 
regarding the fine-tuning and linear probing results, also confirming the competency and robustness of our method.

Concerning the fusion of MIM and CL, significant endeavours \cite{taoSiameseImageModeling2023, yiMaskedImageModeling2022, zhouImageBERTPretraining2021a} are made to enhance pre-training performance on downstream tasks. However, they result in inefficiencies due to increased training iterations, augmented data, auxiliary modules, and diverse loss combinations. For instance, SiameseIM \cite{taoSiameseImageModeling2023}, ConMIM \cite{yiMaskedImageModeling2022} and iBOT \cite{zhouImageBERTPretraining2021a} leverage both hand-crafted data augmentation and more training epochs. Despite our model also integrating masking strategy with contrastive learning, it does not rely on hand-crafted view data augmentations and additional auxiliary modules, reconciling the divergence between efficient visual representation and effective conceptual pre-training. Besides, our model OCL also achieves competitive results on downstream tasks of linear probing and fine-tuning, and more importantly, shows promising scaling behavior.
More comparison results of downstream tasks are provided in the supplementary.

\begin{table}[htb]
    \centering
    \setlength{\tabcolsep}{0.6mm}{
    \begin{tabular}{@{}lcccc@{}}
\toprule
\multicolumn{1}{c}{\multirow{2}{*}{Methods}}            & \multirow{2}{*}{Epochs} & \multicolumn{2}{c}{COCO} & ADE20k \\
\multicolumn{1}{c}{}                                    &                         & $AP^{b}$   & $AP^{m}$   & mIoU   \\ \midrule
\multicolumn{5}{l}{\textbf{ViT-B/16}}                                                                                          \\ \midrule
Supervised   \cite{taoSiameseImageModeling2023}         & 300                     & 47.9       & 42.9        & 47.4   \\
DINO   \cite{caronEmergingPropertiesSelfSupervised2021} & 800                     & 50.1       & 43.4        & 46.8   \\
iBOT    \cite{zhouImageBERTPretraining2021a}            & 1600                    & 51.2       & 44.2        & 50.0   \\
DenseCL   \cite{wangDenseContrastiveLearning2021}       & 400                     & 46.6       & 41.6        & 44.5   \\
MoCo-v3   \cite{chenEmpiricalStudyTraining2021}         & 600                     & 47.9       & 42.7        & 47.3   \\
BEiT \cite{baoBEiTBERTPreTraining2021}                  & 800                     & 49.8       & 44.4        & 47.1   \\
MAE   \cite{heMaskedAutoencodersAre2022}                & 400                     & 50.6       & 45.1        & 45.0   \\
MAE   \cite{heMaskedAutoencodersAre2022}                & 1600                    & 51.6       & 45.9        & 48.1   \\
GreenMIM   \cite{huangGreenHierarchicalVision2022}      & 800                     & 50.0       & 44.1        & -      \\
EsViT   \cite{liEfficientSelfsupervisedVision2021}      & 300                     & -          & -           & 47.3   \\
MixMAE \cite{liuMixMAEMixedMasked2023}                  & 300                     & 52.3       & 46.4        & 49.9   \\
MixMAE \cite{liuMixMAEMixedMasked2023}                  & 600                     & 52.7       & 47.0        & 51.1   \\
SiameseIM   \cite{taoSiameseImageModeling2023}          & 400                     & 50.7       & 44.9        & 49.6   \\
SiameseIM   \cite{taoSiameseImageModeling2023}          & 1600                    & 52.1       & 46.2        & 51.1   \\
SimMIM   \cite{xieSimMIMSimpleFramework2022}            & 300                     & 51.1       & 45.4        & 48.9   \\ \midrule
OCL                                                    & 800                     & 51.5       & 45.5        & 46.1   \\ \midrule
\multicolumn{5}{l}{\textbf{ViT-L/16}}                                                                                          \\ \midrule
\multicolumn{5}{l}{MoCo-v3   \cite{chenEmpiricalStudyTraining2021}}                                                   \\
BEiT \cite{baoBEiTBERTPreTraining2021}                  & 800                     & 53.3       & 47.1        & 53.3   \\
MAE   \cite{heMaskedAutoencodersAre2022}                & 1600`                   & 53.3       & 47.2        & 53.6   \\
SimMIM   \cite{xieSimMIMSimpleFramework2022}            & 800                     & 53.8       & -           & 53.6   \\
MixMAE \cite{liuMixMAEMixedMasked2023}                  & 600                     & 54.3       & 48.2        & 53.8   \\ \midrule
OCL                                                    & 800                     & 53.2       & 47.0        & 53.2   \\ \bottomrule
\end{tabular}
    }
    \caption{\textbf{Comparsion with previous methods on various downstream tasks, including object detection and segmentation on COCO and ADK20K.} We report $\text{AP}^{\text{box}}$ ($\text{AP}^{\text{b}}$) and $\text{AP}^{\text{mask}}$ ($\text{AP}^{\text{m}}$) on COCO, and mIoU on ADE20K. Arch. represents the model architecture, where ViT-B/16 and ViT-L/16 are utilized to validate the performance of various methods.
    }
    \label{table:coco}
\end{table}

Beyond that, we conducted supervised fine-tuning on the COCO dataset for object detection and instance segmentation using the Mask RCNN \cite{heMaskRCNN2017} framework, with our pre-trained encoder serving as the backbone. We follow the setup of MixMAE \cite{liuMixMAEMixedMasked2023}, with the window size to 16 $\times$ 16 to align with the 1024 $\times$ 1024 input image resolution. In terms of semantic segmentation on the ADE20k dataset, We use the UperNet \cite{xiaoUnifiedPerceptualParsing2018} framework with our pre-trained encoder as its backbone, with the changed window size as mentioned above. The results are shown in Table \ref{table:coco}, that our model achieves competitive results compared to other models with efficient pre-training progress. Compared to the contrastive learning method of MoCo-v3, our method obtains 3.6\% improvement of $\text{AP}^{\text{box}}$ under ViT-B/16, especially without view data augmentations. Besides, the consistent performance gains on ADE20K semantic segmentation, achieving +7.1 mIoU improvements from ViT-B to ViT-L, empirically validate our method's scalability across model capacities while maintaining parameter efficiency

\begin{table}[htbp]
    \centering
    \setlength{\tabcolsep}{1mm}{
    \begin{tabular}{@{}lcccc@{}}
    \toprule
    Methods   & Epochs & IN-A & IN-R & IN-S \\ \midrule
    MSN       \cite{assranMaskedSiameseNetworks2022}  & 1200   & 37.5 & 50.0 & 36.3      \\
    iBOT      \cite{zhouImageBERTPretraining2021a}    & 1600   & 42.4 & 50.9 & 36.9      \\
    DenseCL   \cite{wangDenseContrastiveLearning2021} & 400    & 30.8 & 43.8 & 29.9      \\
    MoCo-v3   \cite{chenEmpiricalStudyTraining2021}   & 600    & 32.4 & 49.8 & 35.9      \\
    MAE       \cite{heMaskedAutoencodersAre2022}      & 1600   & 35.9 & 48.3 & 34.5      \\
    SiameseIM \cite{taoSiameseImageModeling2023}      & 1600   & 43.8 & 52.5 & 38.3      \\ \midrule
    OCL      & 800    & 42.2 & 52.3 & 37.6      \\ \bottomrule
    \end{tabular}
    }
    \caption{\textbf{Comparsion with previous methods on generalization capability and robustness on ImageNet-A (IN-A), ImageNet-R (IN-R) and ImageNet-Sketch (IN-S) datasets.}  ViT-B/16 is utilized as the backbone to validate the performance of various methods. Top-1 accuracy is used as the metric.
    }
    \label{table.gen}
\end{table}

Beyond that, we conduct experiments in the domain generalization setting following the SiameseIM \cite{taoSiameseImageModeling2023}, as shown in Table \ref{table.gen}. Experiments are conducted on  ImageNet-A \cite{hendrycksNaturalAdversarialExamples2021}, ImageNet-R \cite{hendrycksManyFacesRobustness2021} and ImageNet-Sketch \cite{wangLearningRobustGlobal2019}, with ImageNet \cite{russakovskyImagenetLargeScale2015} as the source dataset. Our method achieves results with a competitive edge, demonstrating the advantages of our model in generalization and robustness. Notably, our framework achieves competitive performance with significantly fewer training epochs than conventional contrastive approaches, demonstrating remarkable training efficiency while maintaining stable convergence—a critical advantage for scaling to large-scale datasets and complex architectures.

\section{Conclusions and Outlooks}

In this paper, we introduce OCL, a novel, simple, and effective pre-training paradigm for visual conceptual representation. Our approach uses a masking strategy to generate diverse views with fine-grained semantic differences, enabling contrastive learning to classify and learn agreements within a mini-batch. This design eliminates the need for I) addressing semantic conceptual redundancy within images, II) reconstructing images, III) hand-crafted data augmentations, and IV) additional auxiliary modules, thereby improving efficiency and scalability. Experiments showcase the efficiency and scalability of our method, yielding competitive results compared to previous approaches. 
Additionally, ablation experiments provide insights that could inspire future pre-training paradigms.

We hope that our work will inspire future advancements in contrastive pre-training paradigms, specifically efficient visual representation. 
Our further work will focus on enhancing computational efficiency for billion-parameter models without sacrificing representation quality.

\bibliographystyle{unsrt}  
\bibliography{example_paper}  

\newpage
\appendix
\onecolumn

\section{Related Works}

\subsection{Masked Image Modeling}

Inspired by masked language modeling in NLP, the core of masked image modeling is to predict the masked part of the input image. Among them, BEiT  \cite{baoBEiTBERTPreTraining2021} tokenizes image patches through the reconstruction of the individual image using dVAE, and then predicts the tokens of the masked patches to learn visual representation. 
Similarly, MAE  \cite{heMaskedAutoencodersAre2022} utilizes a high masked ratio (75\%) to corrupt the image and directly reconstruct the pixel-level masked image patches. Subsequently, numerous studies have referenced this paradigm for pre-training endeavours, including DropPos  \cite{wangDropPosPreTrainingVision2023}, U-MAE  \cite{zhangHowMaskMatters2022} and CAE \cite{chenContextAutoencoderSelfsupervised2024}. DropPos  \cite{wangDropPosPreTrainingVision2023} incorporates position reconstruction to bolster the spatial awareness of ViTs. U-MAE  \cite{zhangHowMaskMatters2022} introduces a uniformity loss as a regularization to the MAE loss to further encourage the feature consistency of the pre-training, and addresses the dimensional feature collapse. 
CAE \cite{chenContextAutoencoderSelfsupervised2024} decouples the learning processes for image representation and pretext tasks, enabling the pre-trained model to prioritize image representation while disregarding the pretext task. Additionally, following the successful implementation of MAE, its applicability has been extended to diverse disciplines, such as SiamMAE \cite{guptaSiameseMaskedAutoencoders2023} and MR-MAE  \cite{gaoMimicReconstructEnhancing2024}. Similarly, I-JEPA \cite{assranSelfSupervisedLearningImages2023} predicts the feature encoding of the contextual region via MIM, and employs contrastive learning to align the features of neighbouring regions at the feature level.

Besides, latentMIM \cite{wei2025towards} tackles key training challenges in Latent MIM, showcasing its ability to produce spatially diverse, high-level semantic representations.
Context Autoencoders \cite{chen2021empirical} employ an encoder-decoder framework optimized with a combined reconstruction loss and alignment constraint, ensuring predictable representations of missing patches.
data2vec \cite{baevski2022data2vec} predicts representations of missing patches using an online target encoder, eliminating the need for handcrafted augmentations and achieving strong performance across vision, text, and speech modalities. Its successor, data2vec-v2 \cite{baevski2023efficient}, further explores efficient architectures for multimodal learning.

\subsection{Contrastive Learning}

Among these methods, SimCLR \cite{chenSimpleFrameworkContrastive2020} relies on complex pre-processing techniques to create distinct views of an image, aiming to make the task challenging enough to learn effective visual representations during pre-training. However, it requires large batch sizes to generate negative sample pairs, leading to long training times and high computational costs.
DINO \cite{caronEmergingPropertiesSelfSupervised2021} uses a student-teacher architecture to extract visual representations from different views. Similarly, MoCo v3 \cite{chenEmpiricalStudyTraining2021a} employs momentum updates to optimize its auxiliary network. These methods highlight that the core of contrastive learning lies in creating views with significant differences. However, this is challenging due to the inherent semantic redundancy in images.
Additionally, ConCL \cite{yangConCLConceptContrastive2022} generates distinct concepts by cropping images and applies contrastive learning within a teacher-student framework, specifically for pre-training on pathological images.

 DenseCL \cite{wangDenseContrastiveLearning2021} introduces the concept of dense contrastive loss, which calculates the contrastive loss between dense feature vectors generated by the dense projection head at the local feature level, contrasting with traditional contrastive learning methods that operate at the global feature level. Moreover, MaskCo \cite{zhaoSelfSupervisedVisualRepresentations2021} utilizes the teacher-student network for generating image features of query and keys. The key features of different images are used to query certain images to generate positive and negative examples for contrastive learning. Furthermore, MSN \cite{assranMaskedSiameseNetworks2022} leverages data augmentation to create two views known as the anchor view and the target view. Subsequently, a random mask is applied to the anchor view, aligning the representation of the masked anchor view with the clusters of the unmasked target view.

\subsection{Combination between MIM and CL}

There are also a lot of efforts to bridge the gap between mask image modeling and contrastive learning, where the integration of teacher-student models emerges as the prevailing approach. iBOT \cite{zhouImageBERTPretraining2021a}  employs the teacher-student network to independently encode the two augmented views, with the student network processing masked images. The objectives of MIM and CL are jointly trained for self-distillation. Likewise, MST \cite{liMSTMaskedSelfSupervised2021} introduces a masked token strategy leveraging multi-head self-attention maps, which selectively mask the tokens of the student network based on the output self-attention map of the teacher network, ensuring vital foreground remains intact. Similarly, SiameseIM  \cite{taoSiameseImageModeling2023} employs a Siamese network featuring two branches. The online branch encodes the initial view and predicts the representation of the second view based on their relative positions. Meanwhile, the target branch generates the target by encoding the second view. 
MSCN \cite{jingmasked} generates multiple augmented views from input images and applies random masking. These masked views are encoded using a standard ConvNet, with representations optimized through a joint-embedding loss.
RECON \cite{qi2023contrast} integrates generative and contrastive learning paradigms via ensemble distillation, where a generative student model guides a contrastive student to unify both approaches.
CMAE \cite{huang2023contrastive} employs a dual-branch architecture: an online branch with an asymmetric encoder-decoder for reconstructing masked images, and a momentum branch with a momentum encoder for contrastive learning on full images. This design enables holistic feature learning and enhanced discrimination.

ccMIM  \cite{zhangContextualImageMasking2022} leverages a contrastive loss to aid the reconstruction task as a regularizer, facilitating the extraction of image-wide global information from both masked and unmasked patches. Likewise, ConMIM  \cite{yiMaskedImageModeling2022} produces simple intra-image inter-patch contrastive constraints as the sole learning objectives for masked patch prediction, and strengthens the denoising mechanism with asymmetric designs to improve the network pre-training. 
Additionally, CoMAE  \cite{yangCoMAESingleModel2023} also applies CL to assist cross-modal MIM tasks.
Besides, LGP \cite{jiangLayerGraftedPretraining2022} integrates MIM and CL in a sequential cascade manner: early layers are first trained under one MIM loss, on top of which latter layers continue to be trained under another CL loss.

\subsection{Differences of OCL from Existing Methods}

Masking has been adopted as an effective data augmentation technique to enhance training efficiency in several studies \cite{li2023scaling,yang2023attentive,mishra2022simple,wu2022extreme,assran2022masked}. Previous studies have demonstrated its effectiveness in improving model performance and training efficiency \cite{li2023scaling,yang2023attentive}.
ExtreMA \cite{wu2022extreme} employs random masking as a computationally efficient augmentation for Siamese representation learning, accelerating learning and enhancing performance on large datasets. 
MSN \cite{assran2022masked}, the most relevant work to ours, generates two image views: a masked anchor view and an unmasked target view, aiming to cluster their representations. CAN \cite{mishra2022simple} integrates contrastive learning, masked auto-encoding, and diffusion denoising into a unified framework.
Our approach distinguishes itself from previous hybrid methods by achieving superior performance with a better performance-efficiency trade-off. Specifically, our masking strategy generates semantically diverse views and leverages contrastive learning to promote classification agreement within mini-batches. This approach eliminates: (I) semantic redundancy, (II) image reconstruction, (III) hand-crafted augmentations, and (IV) additional auxiliary modules, resulting in enhanced efficiency and scalability.

\section{Experiments}

In this section, we first introduce the utilized datasets for pre-training and various downstream tasks. Subsequently, implementation details are provided. Finally, we present more results on downstream tasks, such as object detection, segmentation and domain generalization.

\subsection{Datasets}

The ImageNet-1k dataset \cite{russakovskyImagenetLargeScale2015}  is a widely used image dataset consisting of 1.28M labeled images across 1k categories, with 50K validation images and 100k test images. The dataset has been instrumental in advancing computer vision research by providing a large-scale benchmark for image classification tasks.  By leveraging the vast amount of labeled images in ImageNet-1k, self-supervised models can learn rich representations of visual data in an unsupervised manner, which can then be fine-tuned on downstream tasks with smaller labeled datasets. Besides, ImageNet-A \cite{hendrycksNaturalAdversarialExamples2021}, ImageNet-R \cite{hendrycksManyFacesRobustness2021} and ImageNet-Sketch \cite{wangLearningRobustGlobal2019} are leveraged for validation of the generalization capability and robustness of the vision model, with the training source of ImageNet.

MSCOCO \cite{linMicrosoftCOCOCommon2014} dataset is a large-scale dataset widely used for object detection and instance segmentation tasks created by Microsoft Research Asia. The COCO Detection dataset contains more than 330K images, including more than 1.5M labeled object instances for 80 different categories. Each object instance is labeled with category, bounding box, and segmentation mask information.

The ADE20K \cite{zhouSceneParsingADE20K2017} semantic segmentation dataset comprises over 20K scene-centric images for 150 semantic categories, meticulously annotated with pixel-level object and object parts labels.

\subsection{Implementation Details}

We employ ViT-Base and ViT-Large as our visual backbones, respectively. Among them, 
Vit-Base consists of 12 transformer encoder layers and an FFN intermediate size of 3,072. The hidden dimensions of the ViT-Base are 768, with 12 attention heads. The number of parameters is about 86 million. The input image size is set to $224\times 224$. In terms of ViT-L/16, ViT-L/16 consists of 24 transformer encoder layers and an FFN intermediate size of 4,096. The input image size is set to $224\times 224$, with a patch size of $16\times 16$. The hidden dimensions of the ViT-Large are 1,024, with 16 attention heads. And, the number of parameters is about 307 million.

In terms of the pre-training progress, the hyperparameters are presented in Table \ref{table:pretrain}. 
We utilize the AdamW optimizer, which is configured with a cosine annealing schedule as the learning policy. The initial base learning rate is set to $1.5\times10^{-4}$, and the AdamW optimizer is employed with hyperparameters $\beta= (0.9, 0.95)$. Additionally, we set the weight decay to 0.05 without dropout. We use the strategy of cosine learning rate decay, with 40 warm-up epochs. Unless otherwise specified, the pre-training of our vision language model consists of 800 epochs, executed on $2\times 2$ NVIDIA A100 GPUs.

\begin{table}[htbp]
    \caption{The pre-training hyperparameters.}
    \label{table:pretrain}
    \centering
    \setlength{\tabcolsep}{3mm}{
        \begin{tabular}{@{}lcc@{}}
        \toprule
                            & ViT-B/16    & ViT-L/16    \\ \midrule
        Training Epochs     & 800         & 800         \\
        Warmup Epochs       & 40          & 40          \\
        Optimizer           & AdamW       & AdamW       \\
        Base Learning Rate  & 1.5e-4      & 1.5e-4      \\
        Learning Rate Decay & Cosine      & Cosine      \\
        Adam $\beta$        & (0.9, 0.95) & (0.9, 0.95) \\
        Weight Decay        & 0.05        & 0.05        \\
        Eff. Batch Size     & 9,600       & 2,400       \\ \bottomrule
        \end{tabular}
        }    
\end{table}

Concerning the downstream tasks of fine-tuning and linear probing on ImageNet, the hyperparameters are shown in Table \ref{table:finetune} and Table \ref{table:lin}.

\begin{table}[htbp]
    \caption{The fine-tuning hyperparameters.}
    \label{table:finetune}
    \centering
    \setlength{\tabcolsep}{3mm}{
        \begin{tabular}{@{}lcc@{}}
        \toprule
                            & ViT-B/16    & ViT-L/16    \\ \midrule
        Training Epochs     & 100         & 50          \\
        Warmup Epochs       & 5           & 5           \\
        Optimizer           & AdamW       & AdamW       \\
        Base Learning Rate  & 5e-4        & 1e-3        \\
        Learning Rate Decay & Cosine      & Cosine      \\
        Adam $\beta$        & (0.9, 0.95) & (0.9, 0.95) \\
        Weight Decay        & 0.05        & 0.05        \\
        Eff. Batch Size     & 1,024       & 1,024       \\ \bottomrule
        \end{tabular}
        }    
\end{table}

\begin{table}[htbp]
    \caption{The linear probing hyperparameters.}
    \label{table:lin}
    \centering
    \setlength{\tabcolsep}{3mm}{    
        \begin{tabular}{@{}lcc@{}}
        \toprule
                            & ViT-B/16 & ViT-L/16 \\ \midrule
        Training Epochs     & 90       & 50       \\
        Warmup Epochs       & 10       & 10       \\
        Optimizer           & LARS     & LARS     \\
        Base Learning Rate  & 0.1      & 0.1      \\
        Learning Rate Decay & Cosine   & Cosine   \\
        Weight Decay        & 0.0      & 0.0      \\
        Eff. Batch Size     & 16,384   & 1,024    \\ \bottomrule
        \end{tabular}
    }
\end{table}

\end{document}